\PassOptionsToPackage{dvipsnames}{xcolor}
\PassOptionsToPackage{table}{xcolor}
\PassOptionsToPackage{numbers, compress}{natbib}
\documentclass{article}
\usepackage[dandb,preprint]{ms}

\usepackage{alltt}
\usepackage[framemethod=tikz]{mdframed}
\usepackage[utf8]{inputenc} 
\usepackage[T1]{fontenc}    
\usepackage{hyperref}       
\usepackage{url}            
\usepackage{booktabs}       
\usepackage{amsfonts}       
\usepackage{nicefrac}       
\usepackage{menukeys}
\usepackage{microtype}      
\usepackage{subcaption}

\usetikzlibrary{decorations.pathreplacing,patterns}
\definecolor{arc0}{RGB}{0, 0, 0}
\definecolor{arc1}{RGB}{30, 147, 255}
\definecolor{arc2}{RGB}{250,61,49}
\definecolor{arc3}{RGB}{0,194,0} 
\definecolor{arc4}{RGB}{255,221,0}
\definecolor{arc5}{RGB}{153,153,153}
\definecolor{arc6}{RGB}{229,59,163}
\definecolor{arc7}{RGB}{255,133,28}
\definecolor{arc8}{RGB}{136,216,241}
\definecolor{arc9}{RGB}{167,37,69} 

\title{ARC-GEN: A Mimetic Procedural Benchmark Generator for the Abstraction and Reasoning Corpus}

\author{%
  Michael D. Moffitt\\
  Google\\
  \texttt{moffitt@google.com}
}

\hypersetup{draft}
\begin{document}

\maketitle

\begin{abstract}
The Abstraction and Reasoning Corpus remains one of the most compelling and challenging benchmarks for tracking progress toward achieving Artificial General Intelligence.  In contrast to other evaluation datasets designed to assess an agent's task-specific skills or accumulated knowledge, the ARC-AGI suite is specifically targeted at measuring {\it skill acquisition efficiency}, a trait that has (so far) been lacking in even the most sophisticated machine learning systems.  For algorithms that require extensive intra-task exemplars, a significant constraint imposed by ARC-AGI is the modest cardinality of its demonstration set, comprising a small number of $\langle${\it input}, {\it output}$\rangle$ grids per task specifying the corresponding transformation.  To embellish the space of viable sample pairs, this paper introduces ARC-GEN, an open-source procedural generator aimed at extending the original ARC-AGI training dataset as faithfully as possible.  Unlike prior efforts, our generator is both {\it exhaustive} (covering all four-hundred tasks) and {\it mimetic} (more closely honoring the distributional properties and characteristics embodied in the initial ARC-AGI-1 release).  We also discuss the use of this generator in establishing a static benchmark suite to verify the correctness of programs submitted to the {\it 2025 Google Code Golf Championship}.
\end{abstract}

\section{Introduction}

The rapid advance of deep learning systems has been accompanied by an incredible explosion in benchmarks used to facilitate their training and subsequent evaluation.  While Large Language Models require inputs tailored to conversation \citep{Chiang2024,Zheng2024} and question answering \citep{White2025}, other systems are designed to ingest data from a wide range of modalities, such as images \citep{Pathak2025,Wu2025}, spoken language \citep{Huang2025,Shah2025}, and long-form videos \citep{Chen2025}.  Recent improvements in special purpose models have also given rise to datasets devoted to certain areas of subject matter expertise, including protein folding \citep{Ye2025}, code generation \& refactoring \citep{Gautam2025,Jain2025,Jimenez2024}, and modern mathematics \citep{Glazer2024}.  Many benchmark suites place a considerable emphasis on {\it knowledge mastery}, containing problems whose solutions often require the expenditure of hours worth of effort on behalf of human experts.  The difficulty of such tasks is perhaps best exemplified by Humanity’s Last Exam \citep{Phan2025}, a closed-ended academic benchmark intended to span the absolute frontier of human knowledge.

For a variety of reasons, one benchmark set in particular -- the {\it Abstraction and Reasoning Corpus} \citep{Chollet2019} -- stands apart from the rest.  First, the dataset is remarkably compact; i.e., the entire suite of training instances requires only 1.44MB of storage (uncompressed), small enough to fit onto a 3.5" floppy disk.\footnote{By comparison, the ImageNet Object Localization dataset \citep{Howard2018} is 167.62 GB, roughly 100,000$\times$ larger.}  Second, the problems are easily approachable (if not trivial) for most humans, with the vast majority capable of being solved by non-experts \citep{LeGris2024} and even children \citep{Opielka2024}.  Third, the benchmark has proven somewhat resistant to deep learning techniques, as partly evidenced by the (unclaimed) top prize offered for a sufficiently high-scoring submission to the \$1 million ARC Prize 2024 challenge \citep{arc-prize-2024,Chollet2024}.

One key characteristic of this dataset (hereafter referred to as ARC-AGI-1)\footnote{A successor to this set (named ARC-AGI-2) was released on March 24\textsuperscript{th}, 2025 for ARC Prize 2025 \citep{Chollet2025,arc-prize-2025}.} is the limited cardinality of samples per task, which are represented as two-dimensional arrays of digits.  Since the ARC-AGI format embodies transformations in a way that is highly differentiated from other problem representations available on the internet, this lack of examples can pose a challenge when training a model how to identify the appropriate mapping for any given task. Likewise, the risk of contamination grows significantly as ARC-related tasks increase in popularity, thus complicating the fair evaluation of ARC solvers.  These issues have led to the development of various ARC-themed benchmark generators, yet such systems are not designed (and therefore certainly not guaranteed) to represent all original puzzles exactly, or ensure compatibility with programs that have been developed to solve them \citep{Hodel2024,Li2025}.

In this paper, we present an open-source procedural benchmark generator named ARC-GEN aimed at extending the original ARC-AGI training dataset as faithfully as possible.  Unlike previous efforts, our generator is both {\it exhaustive} (covering all four-hundred tasks) and {\it mimetic} (honoring distributional properties and characteristics similar to those embodied in the initial ARC-AGI-1 release).  We demonstrate these attributes empirically by validating its ability to reproduce all examples in the original benchmark suite, and by verifying its consistency with respect to programs designed to implement each ARC-AGI transformation. Finally, we discuss an application of this generator in establishing a test harness for the {\it 2025 Google Code Golf Championship}, a competition that requires a large number of examples to certify the correctness and generality of submitted programs.

\section{Background}

We begin by reviewing some basic concepts surrounding abstraction and reasoning, along with their manifestation in the ARC benchmark suite.  Our terminology is consistent with prior works studying various measures of intelligence \citep{Chollet2019}, albeit truncated for the sake of expositional brevity.

\subsection{Algorithmic Information Theory}

Consider a list of tasks $\langle \mathcal{T}_1, \mathcal{T}_2, ...\,, \mathcal{T}_n \rangle$ where each task $\mathcal{T}_i$ performs some arbitrary predefined state transformation:
$$\mathcal{T}_i: \mathcal{S}_i \rightarrow \mathcal{S}_i' \;\; \forall i \in [1, n]$$

In addition, suppose a finite set of examples $\langle \mathcal{P}_1, \mathcal{P}_2, ...\,, \mathcal{P}_n \rangle$ such that each example in $\mathcal{P}_i$ is consistent with the corresponding transformation $\mathcal{T}_i$:

$$\mathcal{P}_i: \bigl \langle (s_{i,1},s_{i,1}'), (s_{i,2},s_{i,2}'), ...\,\bigr \rangle \;\;|\;\; s_{i,j} \in \mathcal{S}_i, \,s_{i,j}' = \mathcal{T}_i(s_{i,j})$$

The meta-task of {\it skill acquisition} is to infer $\mathcal{T}_i$ from the elements of $\mathcal{P}_i$ alone such that for any new state $\widehat{s}_{i,j} \in \mathcal{S}_i$, the expected output $\widehat{s}_{i,j}' = \mathcal{T}_i(\widehat{s}_{i,j})$ can be produced.  This process requires {\it generalization}, and the intelligence of a system can be argued to correlate directly with its ability to perform this generalization {\it efficiently}.

\subsection{The Abstraction and Reasoning Corpus}

First introduced in 2019, the Abstraction and Reasoning Corpus \citep{Chollet2019} considers a variety of tasks involving the transformation of two-dimensional grids.  Such puzzles usually have a clear visual or geometric interpretation (with common themes including {\it translation}, {\it rotation}, {\it dilation}, {\it submatrix selection}, etc.), yet no two tasks are exactly the same. Most lend themselves to concise natural language descriptions \citep{Acquaviva2022}; refer to Figure \ref{fig:task125_description} for an English translation of one puzzle (ID: {\small {\tt 543a7ed5}}).

\begin{figure}[t]
  \centering
  \resizebox{0.9\textwidth}{!}{\input{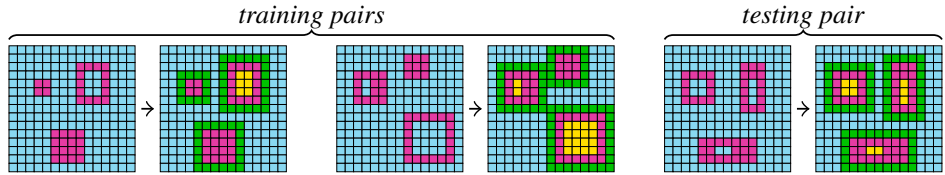}}\vspace{-0.05in}
  \caption{All examples in the original ARC-AGI-1 benchmark suite for puzzle ID {\small {\tt 543a7ed5}}.}
  \label{fig:task125_arc_agi}
\end{figure}

The number of examples varies by problem, but typically ranges between three and five per task.   For illustration, we show in Figure \ref{fig:task125_arc_agi} the only $\langle${\it input}, {\it output}$\rangle$ pairs provided for puzzle {\small {\tt 543a7ed5}}.  This relatively small number of examples is reasonably justified by an expectation that any competent agent should need only a handful such instances to produced the desired transformation outputs, and (thankfully) this few-shot behavior has commonly been observed in practice \citep{Brown2020,Ravi2017}.  A secondary rationale is the enormous amount of manual work that went into producing \& verifying this set, as all examples were painstakingly constructed by hand over the course of several months.

\begin{figure}[t]
\mdfsetup{%
middlelinecolor=gray,
middlelinewidth=1pt,
backgroundcolor=gray!5,
roundcorner=3pt}
\begin{mdframed}
\footnotesize
\begin{alltt}
[INPUT] Three non-overlapping pink rectangles (potentially hollowed out) placed
upon a 15x15 cyan background.

[OUTPUT] Those same three pink rectangles each surrounded by a 1-pixel green
border and all holes shaded yellow.
\end{alltt}
\end{mdframed}\vspace{-0.1in}
\caption{A natural language description of an ARC-AGI-1 puzzle (ID: {\small {\tt 543a7ed5}}).}
\label{fig:task125_description}
\end{figure}

\subsection{ARC Solvers}

Despite the apparent simplicity of the ARC formulation, a surprisingly wide variety of approaches have emerged in an attempt to address it.  Many efforts have focused on program-oriented solutions -- e.g., {\it program search} \citep{Alford2022,Macfarlane2025}, {\it program synthesis} \citep{Ainooson2023,Bober2024,Hocquette2025,Huang2023,Witt2025}, {\it program sampling} \citep{Butt2024}, {\it program repair} \citep{Tang2024repair}, and {\it program induction} \citep{Ouellette2024,Wang2024} -- all of which involve the construction of task-specific source codes capable of transforming any subsequent input image.  An alternative technique involves the use of variational autoencoders to convert examples into low-dimensional latent vectors, allowing the direct production of output images by applying vector arithmetic \citep{Veldkamp2023}.  Given that the patterns embodied in most puzzles tend to be simple, the lossless information compression proposed in CompressARC has also demonstrated potential (despite the absence of pretraining) \citep{Liao2025}, and since certain tasks are more amenable to some techniques than others, a recent implementation considers an ensemble method to exploit their respective puzzle-specific strengths \citep{Li2025}.  Finally, both Large Language Models \citep{Berman2024,Greenblatt2024,Tan2023} and small Transformer models \citep{FletcherHill2024,Puget2024} have shown exceptional promise, with newer algorithms employing {\it test-time training} \citep{Akyurek2025} and the {\it language of thought hypothesis} \citep{Lee2025} to alleviate the shortcomings of these neural architectures.

\subsection{ARC-Related Datasets and Generators}

\begin{figure}[t]
  \centering
  \resizebox{1.0\textwidth}{!}{\input{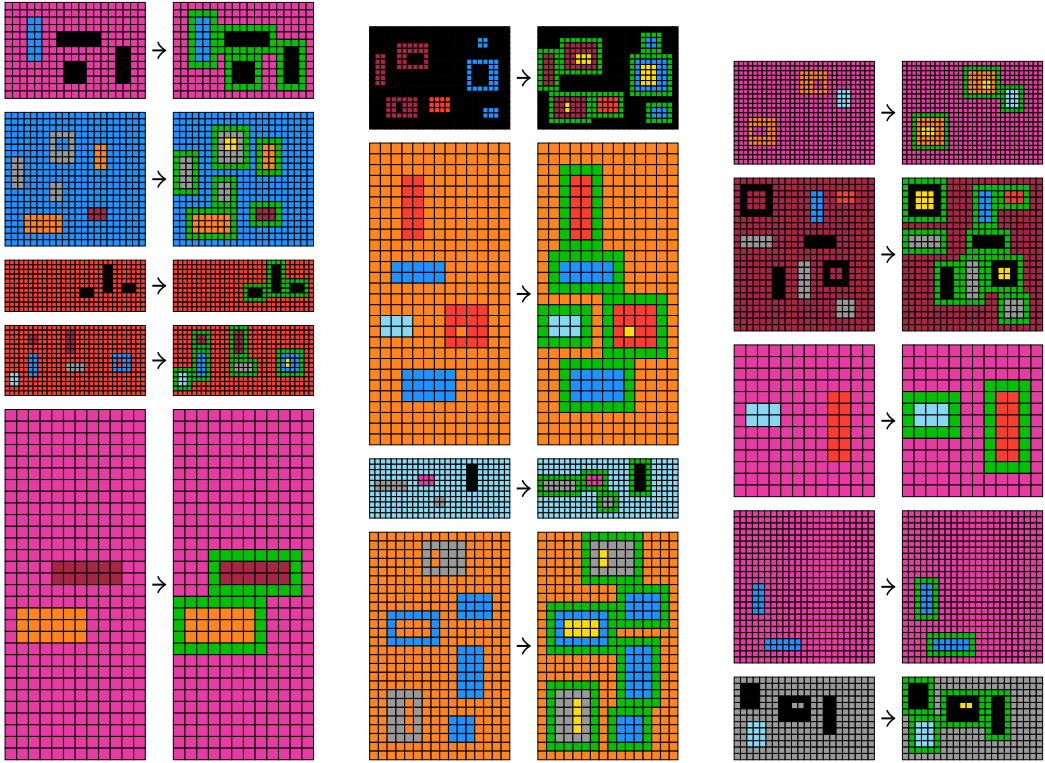}}\vspace{-0.05in}
  \caption{Examples for puzzle ID {\small {\tt 543a7ed5}} produced by the RE-ARC procedural generator.}
  \label{fig:task125_re_arc}
\end{figure}

A common theme across many ARC solvers is the use of {\it data augmentation} \citep{Franzen2024} to produce $\langle${\it input}, {\it output}$\rangle$ pairs beyond those originally released by the ARC Prize Foundation.  One such effort named ConceptARC \citep{Moskvichev2023} introduced a suite of new tasks formed around sixteen specific {\it concept groups} (e.g., sameness, extraction, counting, etc.). In addition to static problem sets, a handful of (unofficial) procedural generators have been developed to further embellish the sample space.  For instance, the BARC implementation relies on synthetic generation using {\it seeds} (each corresponding to one of the original puzzles) which are then mutated and recombined to produce many thousands of variations.  Although the instances produced by these seed generators are intended to reflect those in the official set, only a fraction of tasks are represented, and nearly a third have been shown to produce incorrect results.  Finally, the RE-ARC project \citep{Hodel2024} offers complete task coverage -- providing a DSL-based generator for each of the four-hundred training puzzles -- but lifts various distributional constraints to increase the diversity of the sample space.\footnote{\url{https://github.com/michaelhodel/arc-dsl/}}  For illustration, refer to Figure \ref{fig:task125_re_arc} in which several additional degrees of freedom (e.g., the number of boxes, their colors, grid dimensions) have been introduced.  Using the nomenclature of Algorithmic Information Theory, one could say that for every original task $\mathcal{T}_i: \mathcal{S}_i \rightarrow \mathcal{S}_i'$, the RE-ARC generator draws its examples from an extended task $\overline{\mathcal{T}}_i$:
$$\overline{\mathcal{T}}_i: \overline{\mathcal{S}}_i \rightarrow \overline{\mathcal{S}}_i' \;\;|\;\; \overline{\mathcal{S}}_i \supseteq \mathcal{S}_i, \,\overline{\mathcal{T}}_i(s_{i,j}) = \mathcal{T}_i(s_{i,j})$$
Depending on the context, a broader sampling space can offer significant benefits -- most notably, allowing the construction and validation of systems that exhibit much stronger generalization capabilities -- but it is considerably less useful when evaluating systems explicitly designed to solve the original set of tasks (such as the motivating application we consider in Section \S \ref{code_golf}).

\section{The ARC-GEN Procedural Benchmark Generator}

The focus of our work is the procedural generation of benchmarks that complement the ARC-AGI-1 dataset, subject to two important design criteria.  The first is that we seek to construct a generator that is {\it exhaustive} -- that is, covering the complete array of all four-hundred training tasks -- which (given the wide range of core knowledge priors expressed in these transformations) is a rather significant undertaking.  The second is that we wish for this generator to be {\it mimetic}, emulating virtually the same constraints and distributional properties that are present in the original suite.  The result of these efforts is ARC-GEN, an open-source library that is freely available for both academic and commercial use:
\begin{center}
  \url{https://github.com/google/ARC-GEN}
\end{center}
In the following sections, we outline the structure of a typical sub-generator, illustrating the key features and components that differentiate our approach from previous works.

\subsection{Parameterization}

For each task, our generator defines a parameterized {\small {\tt generate()}} function whose arguments dictate the abstract entities contained in an example.  We show in Figure \ref{fig:task_125_generator} the function for puzzle {\small {\tt 543a7ed5}}, which accepts parameters describing the locations, dimensions, and colors of all boxes.  These variables are all task-specific; common attributes for other puzzles include pixel coordinates, ``sprites'' (small bitmap objects), and boolean indicators to toggle certain grid manipulations such as reflection or transposition.

By default, most parameterization arguments are unspecified (using the constant {\small {\tt None}}), in which case the generator will automatically populate their values using random numbers.  Special care must be taken to ensure that these assignments are consistent with the task -- for instance, that boxes do not overlap -- and so it is not unusual to find conditional logic in this section dedicated to checking such constraints and (in the case of violation) attempting different combinations of parameter values.

\begin{figure}[t]
\mdfsetup{%
middlelinecolor=gray,
middlelinewidth=1pt,
backgroundcolor=gray!5,
roundcorner=3pt}
\begin{mdframed}
\begin{footnotesize}
\begin{alltt}\input{task125_generator}\end{alltt}
\end{footnotesize}
\end{mdframed}\vspace{-0.1in}
\caption{Our procedural generation code for just one of the four-hundred puzzles (ID: {\small {\tt 543a7ed5}}).}
\label{fig:task_125_generator}
\end{figure}

\subsection{Generation}

Unlike prior procedural generators for ARC, our core generation logic is decoupled from parameterization, allowing both the {\it validation} and optional {\it variation} of tasks (as explained in subsequent sections).  The purpose of this generation stanza is to translate all high-level task-specific arguments -- whether specified as function parameters or determined synthetically -- into pixel-perfect representations of the input and output grids.

\begin{figure}[t]
  \centering
  \resizebox{1.0\textwidth}{!}{\input{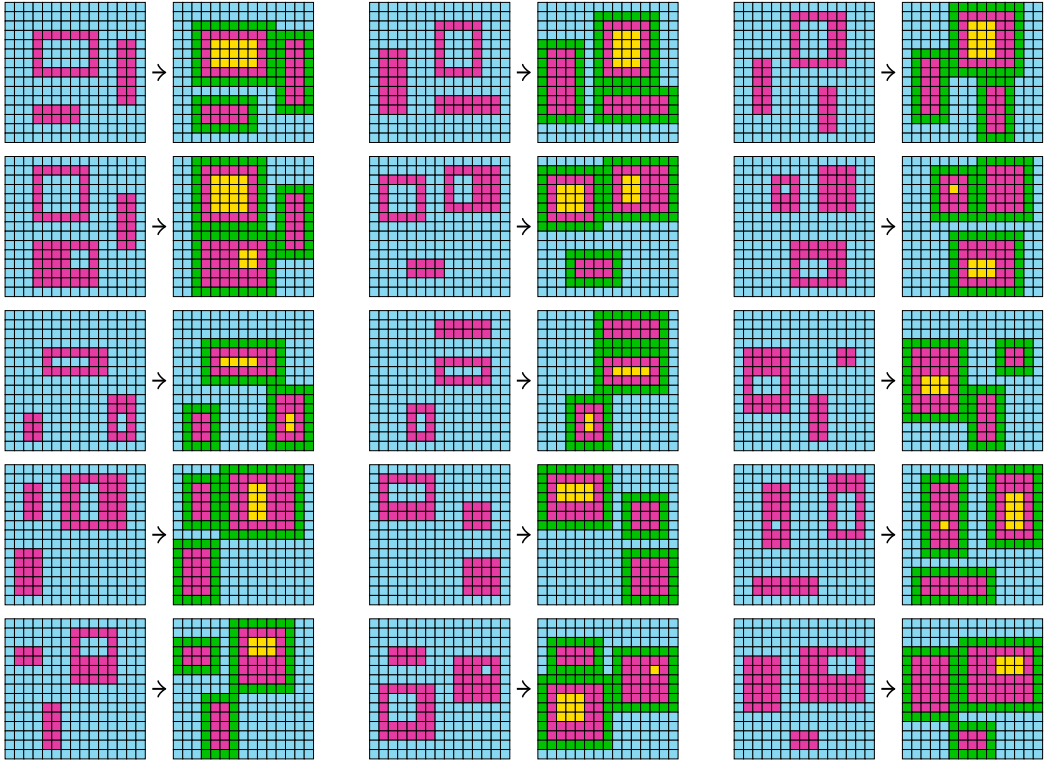}}\vspace{-0.05in}
  \caption{Examples for puzzle ID {\small {\tt 543a7ed5}} produced by our ARC-GEN procedural generator.}
  \label{fig:task125_arc_gen}
\end{figure}

In the case of puzzle {\small {\tt 543a7ed5}}, the process is quite simple: we iterate over the set of boxes, and then through all coordinates that each box occupies (marking the appropriate grid cells with the corresponding color); refer to Figure \ref{fig:task125_arc_gen} for an array of such examples.  For most cases, we have found that the effort required to code the generator for a task is much more straightforward than that required to produce its solution, since we maintain full observability into the output grid and can mask it as necessary in order to create its input in tandem.

\subsection{Validation}

The design of any procedural benchmark generator carries the risk of {\it under-specification}, where the original task $\mathcal{T}_i$ has been inadvertently reduced to one that operates across a narrower range of inputs:
$$\widetilde{\mathcal{T}}_i: \widetilde{\mathcal{S}}_i \rightarrow \widetilde{\mathcal{S}}_i' \;\;|\;\; \widetilde{\mathcal{S}}_i \subseteq \mathcal{S}_i, \,\widetilde{\mathcal{T}}_i(\widetilde{s}_{i,j}) = \mathcal{T}_i(\widetilde{s}_{i,j})$$

Returning to our running example, it would be much easier to implement a generator that strictly produces {\it closed} boxes, forgoing the additional logic required to properly size and place the occasional holes. However, there is only marginal utility in such a tool, as it might satisfy a simplified transformation that (either accidentally or deliberately) fails when exposed to a wider range of examples drawn from the original sampling space.

In order to demonstrate the behavioral completeness of our library, we include a {\small {\tt validate()}} function for each task $\mathcal{T}_i$ that contains the appropriate parameters to reproduce the {\it entire sequence} of training and test pairs provided in ARC-AGI-1:

$$\mathcal{V}_i(\mathcal{T}_i, \alpha_i) = \mathcal{P}_i = \bigl \langle (s_{i,1},s_{i,1}'), (s_{i,2},s_{i,2}'), ...\,\bigr \rangle$$

These validation parameters $\alpha_i$ (all derived manually) also serve as functional unit tests for anyone wishing to submit modifications or improvements to our open-source library, which are always welcome.

\subsection{Variations}

\begin{figure}[t]
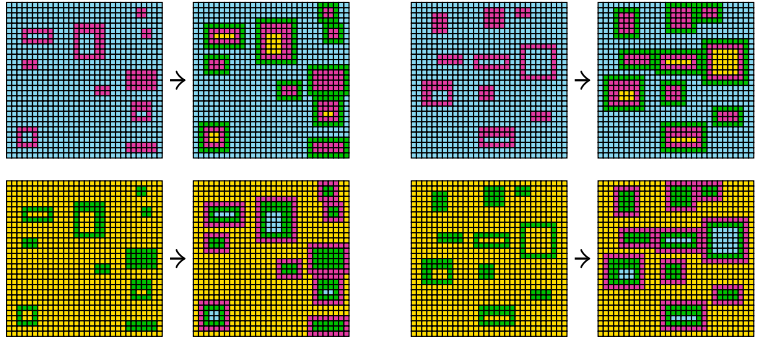

  \centering
  \begin{subfigure}{1.0\textwidth}
  \centering
  \resizebox{0.725\textwidth}{!}{\input{task125_mega}}\vspace{-0.05in}
  \end{subfigure}\vspace{0.15in}
  \begin{subfigure}{1.0\textwidth}
  \centering
  \resizebox{0.725\textwidth}{!}{\input{task125_mega_inverted}}\vspace{-0.05in}
  \end{subfigure}
  \caption{``Large'' and ``inverted'' variations on puzzle ID {\small {\tt 543a7ed5}} produced by ARC-GEN.}
  \label{fig:task125_mega}
\end{figure}

As noted earlier, there is significant value in employing procedural generators to expand the diversity of examples used to train and evaluate new ARC solvers.  In ARC-GEN, we allow such customization for nearly every task, delegating control of the desired variation(s) to callees of our library.  We display four such examples in Figure \ref{fig:task125_mega} for which the values of {\small {\tt boxes}} \& {\small {\tt size}} have been increased, along with extra overrides to introduce color variations.

\section{Evaluation}
\label{evaluation}

In Table \ref{tab:comparison}, we provide a comparison of ARC-GEN vs. the two predominant ARC-flavored generation libraries; specifically, the seeds present in BARC's synthetic generator \citep{Li2025} and the generation utilities in RE-ARC \citep{Hodel2024}.  For each system, we report its task coverage across the ARC-AGI-1 training puzzles, as well as its validation coverage (which, at present, is provided only by ARC-GEN).  Since both BARC and RE-ARC include implementations for the desired transformations (called {\it source programs} and {\it verifiers}, respectively), we report the success rate of these programs across the samples produced by each generator.  Unsurprisingly, the distributional variance in RE-ARC causes problems for most BARC programs, and the low task representation in BARC (coupled with occasional incompatibility) leads to relatively few successes against the verifiers included with RE-ARC.  In contrast, the success rate of ARC-GEN is 100\% across all implementations.

\begin{table}[b]
  \caption{A comparison of procedural benchmark generators for the ARC-AGI-1 dataset.}
  \label{tab:comparison}
  \small
  \centering
  \begin{tabular}{r|ccc}
    \toprule
             & {\bf Examples from} & {\bf Examples from} & {\bf Examples from} \\
             & {\bf seeds in the BARC} & {\bf the RE-ARC} & {\bf our ARC-GEN} \\
             & {\bf Generator} \citep{Li2025} & {\bf Generator} \citep{Hodel2024} & {\bf Generator} \\
    \midrule
                  {\bf Task Coverage} & \textcolor{red}{27\%} & 100\% & \textcolor{ForestGreen}{100\%}\\
            {\bf Validation Coverage} &  \textcolor{gray}{N/A} &   \textcolor{gray}{N/A} & \textcolor{ForestGreen}{100\%}\\
      {\bf BARC Source Success Rate} &  100\% &  \textcolor{red}{3.7\%} & \textcolor{ForestGreen}{100\%}\\
    {\bf RE-ARC Verifier Success Rate} &  \textcolor{red}{16.25\%} &   100\% & \textcolor{ForestGreen}{100\%}\\
    \bottomrule
  \end{tabular}
\end{table}

\section{An Application to the 2025 Google Code Golf Championship}
\label{code_golf}

The aforementioned approach of {\it program synthesis} is a particularly compelling attack vector to solve ARC-AGI, due in part to the many recent advances in automated code generation \citep{Austin2021,Bowers2023,Chen2021,Devlin2017,Du2024,Ellis2021,Evans2021,Fried2023,Gao2023,Li2024icml,Li2024,Li2022,Nijkamp2023,Qiu2025,Schmidhuber2004,Tang2023,Tang2024worldcoder}.  Yet, aside from DSL-based solutions that defer the majority of grid manipulations to a library of specialized subroutines, there exists no canonical set of reference programs (to our knowledge) that can aid in the training and fine-tuning of ARC-oriented solvers.

\begin{figure}[t]
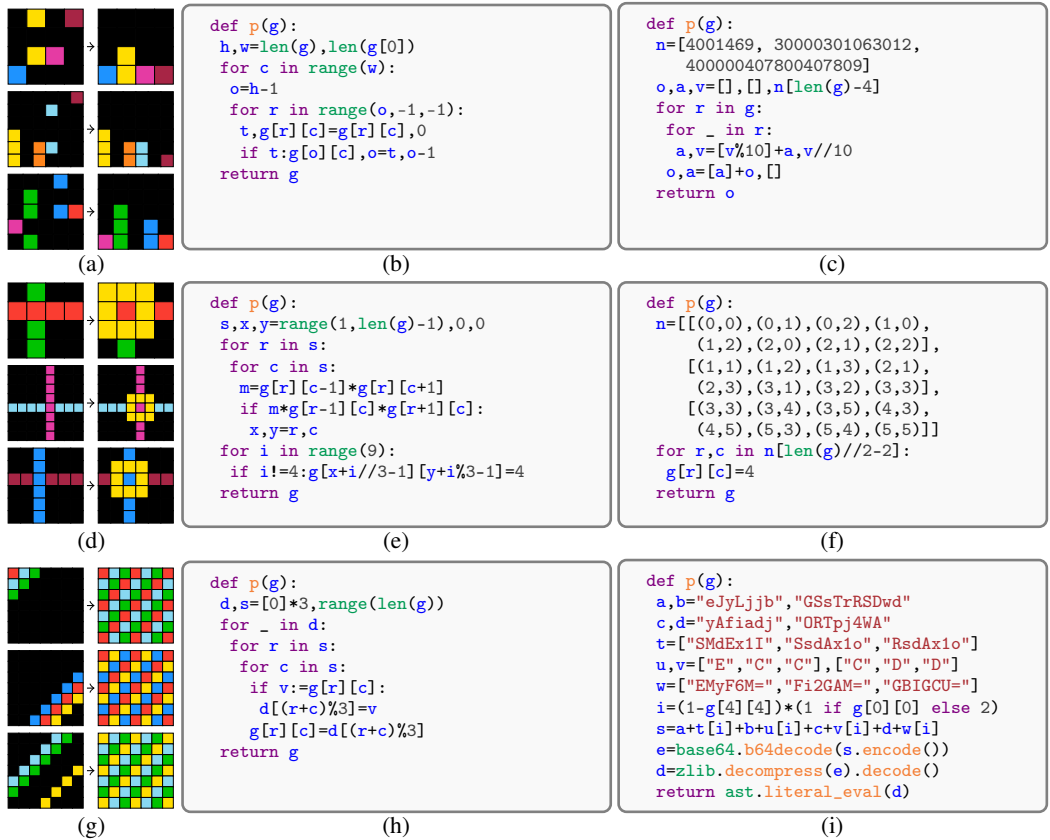

\centering

\begin{subfigure}{0.16\textwidth}\resizebox{1.025\textwidth}{!}{\input{task032_arc_agi}}\vspace{-0.07in}\caption{}\end{subfigure}\hfil
\begin{subfigure}{0.41\textwidth}
\mdfsetup{middlelinecolor=gray,middlelinewidth=1pt,backgroundcolor=gray!5,roundcorner=3pt}
\begin{mdframed}\scriptsize \begin{alltt}\input{task032_code_minimal}\vspace{0.28in}\end{alltt}\end{mdframed}\vspace{-0.13in}\caption{}\end{subfigure}\hfil
\begin{subfigure}{0.41\textwidth}
\mdfsetup{middlelinecolor=gray,middlelinewidth=1pt,backgroundcolor=gray!5,roundcorner=3pt}
\begin{mdframed}\scriptsize \begin{alltt}\input{task032_code_compressed}\vspace{0.20in}\end{alltt}\end{mdframed}\vspace{-0.13in}\caption{}\end{subfigure}

\begin{subfigure}{0.16\textwidth}\resizebox{1.025\textwidth}{!}{\input{task151_arc_agi}}\vspace{-0.05in}\caption{}\end{subfigure}\hfil
\begin{subfigure}{0.41\textwidth}
\mdfsetup{middlelinecolor=gray,middlelinewidth=1pt,backgroundcolor=gray!5,roundcorner=3pt}
\begin{mdframed}\scriptsize \begin{alltt}\input{task151_code_minimal}\vspace{0.06in}\end{alltt}\end{mdframed}\vspace{-0.13in}\caption{}\end{subfigure}\hfil
\begin{subfigure}{0.41\textwidth}
\mdfsetup{middlelinecolor=gray,middlelinewidth=1pt,backgroundcolor=gray!5,roundcorner=3pt}
\begin{mdframed}\scriptsize \begin{alltt}\input{task151_code_compressed}\vspace{0.06in}\end{alltt}\end{mdframed}\vspace{-0.13in}\caption{}\end{subfigure}

\begin{subfigure}{0.16\textwidth}\resizebox{1.025\textwidth}{!}{\input{task007_arc_agi}}\vspace{-0.05in}\caption{}\end{subfigure}\hfil
\begin{subfigure}{0.41\textwidth}
\mdfsetup{middlelinecolor=gray,middlelinewidth=1pt,backgroundcolor=gray!5,roundcorner=3pt}
\begin{mdframed}\scriptsize \begin{alltt}\input{task007_code_minimal}\vspace{0.20in}\end{alltt}\end{mdframed}\vspace{-0.13in}\caption{}\end{subfigure}\hfil
\begin{subfigure}{0.41\textwidth}
\mdfsetup{middlelinecolor=gray,middlelinewidth=1pt,backgroundcolor=gray!5,roundcorner=3pt}
\begin{mdframed}\scriptsize \begin{alltt}\input{task007_code_compressed}\end{alltt}\end{mdframed}\vspace{-0.13in}\caption{}\end{subfigure}

\caption{Examples, near-minimal programs, and hardcoded solutions for three ARC tasks.}
\label{fig:code_golf}
\end{figure}

To help curate a community-owned collection of pure Pythonic solutions for these tasks, the recent {\it 2025 Google Code Golf Championship} invited participants from around the world to contribute functional implementations that solve each of the four-hundred ARC-AGI-1 training puzzles.\footnote{\url{https://www.kaggle.com/competitions/google-code-golf-2025}} In order to gamify this process while soliciting the shortest possible implementations, submissions were scored by {\it program length} -- for example, we show in Figure \ref{fig:code_golf} three such source codes (b,e,h) that, while not necessarily minimal, are at least very close to it.  Regrettably, our experience administering similar contests suggested that many submissions might have instead resembled the rightmost programs where outputs are hardcoded in various ways, e.g. storing grid values in large integers (c), sparse lists (f), and compressed strings (i).  There are an unbounded number of strategies to represent these grids, so it is difficult to disincentivize such submissions if restricting verification to the original ARC-AGI-1 examples.

To prevent overfitting, we thus employed ARC-GEN to synthesize {\it hundreds} of examples per task (totaling 100,000 samples in all), requiring each submission to produce correct outputs across every image pair to be considered for eligibility.\footnote{\url{https://www.kaggle.com/datasets/arcgen100k/the-arc-gen-100k-dataset}}  Although it may still be possible to configure hardcoded outputs for individual inputs, such programs are unlikely to be competitive due to their extreme character count.

\begin{figure}[t]
\centering
\begin{subfigure}{\textwidth}\resizebox{1.0\textwidth}{!}{\input{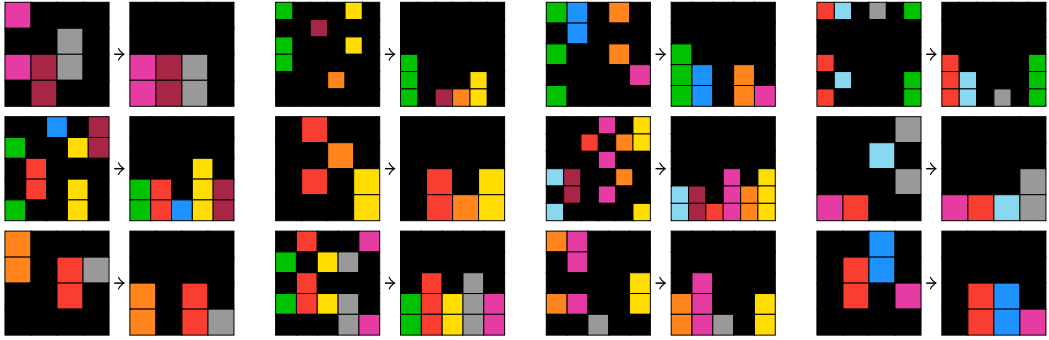}}
  \caption{A subset of ARC-GEN examples for puzzle ID {\small {\tt 1e0a9b12}}.}
\end{subfigure}\vspace{0.25in}
\begin{subfigure}{\textwidth}\resizebox{1.0\textwidth}{!}{\input{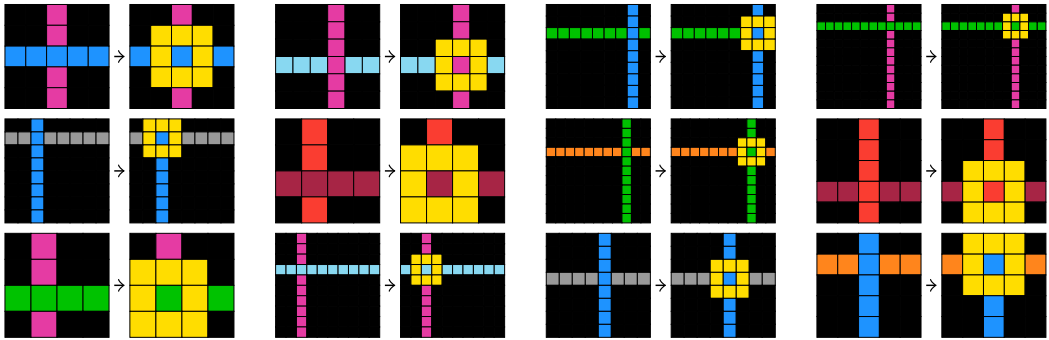}}
  \caption{A subset of ARC-GEN examples for puzzle ID {\small {\tt 67a423a3}}.}
\end{subfigure}\vspace{0.25in}
\begin{subfigure}{\textwidth}\resizebox{1.0\textwidth}{!}{\input{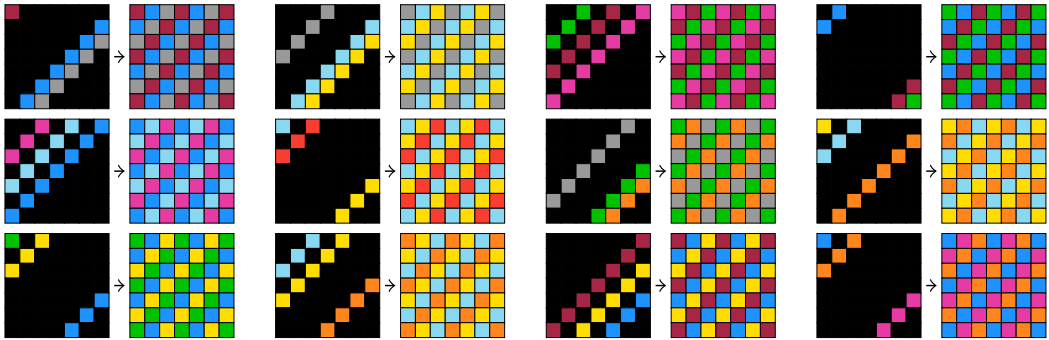}}
  \caption{A subset of ARC-GEN examples for puzzle ID {\small {\tt 05269061}}.}
\end{subfigure}\vspace{0.25in}
\caption{Examples for various puzzles produced by our ARC-GEN procedural generator.}
\label{fig:code_golf_arc_gen}
\end{figure}

\section{Limitations}
\label{limitations}

It is important to note that previous efforts target a different design criteria than we do -- specifically, amplifying task diversity -- which is difficult to capture quantitatively and not reflected in our results.  A more apt comparison for that objective might be the efficacy of ARC solvers trained on these samples, which we defer to future studies (as it lies outside the scope of our work).  The assessment of these generators is also highly dependent on the availability of functional source programs.  Ideally, we would have an broader array of both correct and incorrect programs, which (although currently in short supply) may become more plentiful following the conclusion of our code golf competition. In addition, the limitations of ARC-GEN are tightly intertwined with the spectrum of concepts embodied by ARC-AGI-1.  A wider variety of capabilities (i.e., symbolic interpretation, compositional reasoning, and contextual rule application) are emphasized in ARC-AGI-2 \citep{Chollet2025}, and will eventually require the development of an entirely new class of generators.  Finally, early previews of ARC-AGI-3 showcase an interactive action-oriented sample space, thus involving variations of {\it automata synthesis} \citep{Das2023} where generators (albeit significantly more complicated ones) are likely to play a critical role.\footnote{\url{https://arcprize.org/arc-agi/3/}}

\section{Conclusion}

The ARC-GEN procedural generator presented in this work aims to address the twin goals of {\it maximizing task coverage} while simultaneously {\it minimizing distributional deviation} with respect to the seminal ARC-AGI-1 dataset.  Each sub-generator included in our open-source release is not only capable of producing new mimetic examples, but also parameterized in such a way that allows the complete reproduction of the original benchmark suite, as well as the creation of a broader variety of tasks.  Our library has many potential uses, including (but not limited to) the verification of programs designed to generalize beyond the specific examples included in the original ARC-AGI-1 distribution.  We look forward to seeing the use of our tool, along with others, in the pursuit of systems devoted to tackling Artificial General Intelligence.

\interlinepenalty=10000

\begin{thebibliography}{67}
\providecommand{\natexlab}[1]{#1}
\providecommand{\url}[1]{\texttt{#1}}
\expandafter\ifx\csname urlstyle\endcsname\relax
  \providecommand{\doi}[1]{doi: #1}\else
  \providecommand{\doi}{doi: \begingroup \urlstyle{rm}\Url}\fi

\bibitem[Chiang et~al.(2024)Chiang, Zheng, Sheng, Angelopoulos, Li, Li, Zhu, Zhang, Jordan, Gonzalez, and Stoica]{Chiang2024}
Wei-Lin Chiang, Lianmin Zheng, Ying Sheng, Anastasios~Nikolas Angelopoulos, Tianle Li, Dacheng Li, Banghua Zhu, Hao Zhang, Michael~I. Jordan, Joseph~E. Gonzalez, and Ion Stoica.
\newblock Chatbot {A}rena: {A}n {O}pen {P}latform for {E}valuating {LLM}s by {H}uman {P}reference.
\newblock In \emph{Proceedings of the 41st International Conference on Machine Learning}, ICML 2024, pages 8359--8388, 2024.
\newblock URL \url{https://proceedings.mlr.press/v235/chiang24b.html}.

\bibitem[Zheng et~al.(2024)Zheng, Chiang, Sheng, Li, Zhuang, Wu, Zhuang, Li, Lin, Xing, Gonzalez, Stoica, and Zhang]{Zheng2024}
Lianmin Zheng, Wei-Lin Chiang, Ying Sheng, Tianle Li, Siyuan Zhuang, Zhanghao Wu, Yonghao Zhuang, Zhuohan Li, Zi~Lin, Eric Xing, Joseph~E. Gonzalez, Ion Stoica, and Hao Zhang.
\newblock {LMSYS}-{C}hat-1{M}: {A} {L}arge-{S}cale {R}eal-{W}orld {LLM} {C}onversation {D}ataset.
\newblock In \emph{Proceedings of the 12th International Conference on Learning Representations}, ICLR 2024, 2024.
\newblock URL \url{https://openreview.net/forum?id=BOfDKxfwt0}.

\bibitem[White et~al.(2025)White, Dooley, Roberts, Pal, Feuer, Jain, Shwartz-Ziv, Jain, Saifullah, Dey, Shubh-Agrawal, Sandha, Naidu, Hegde, LeCun, Goldstein, Neiswanger, and Goldblum]{White2025}
Colin White, Samuel Dooley, Manley Roberts, Arka Pal, Benjamin Feuer, Siddhartha Jain, Ravid Shwartz-Ziv, Neel Jain, Khalid Saifullah, Sreemanti Dey, Shubh-Agrawal, Sandeep~Singh Sandha, Siddartha~Venkat Naidu, Chinmay Hegde, Yann LeCun, Tom Goldstein, Willie Neiswanger, and Micah Goldblum.
\newblock Live{B}ench: {A} {C}hallenging, {C}ontamination-{L}imited {LLM} {B}enchmark.
\newblock In \emph{Proceedings of the 13th International Conference on Learning Representations}, ICLR 2025, 2025.
\newblock URL \url{https://openreview.net/forum?id=sKYHBTAxVa}.

\bibitem[Pathak et~al.(2025)Pathak, Marjit, Vyas, and Rawat]{Pathak2025}
Priyank Pathak, Shyam Marjit, Shruti Vyas, and Yogesh~S Rawat.
\newblock {LR}0.{FM}: {L}ow-{R}esolution {Z}ero-{S}hot {C}lassification {B}enchmark for {F}oundation {M}odels.
\newblock In \emph{Proceedings of the 13th International Conference on Learning Representations}, ICLR 2025, 2025.
\newblock URL \url{https://openreview.net/forum?id=AsFxRSLtqR}.

\bibitem[Wu et~al.(2025)Wu, Biamby, Quenum, Gupta, Gonzalez, Darrell, and Chan]{Wu2025}
Tsung-Han Wu, Giscard Biamby, Jerome Quenum, Ritwik Gupta, Joseph~E. Gonzalez, Trevor Darrell, and David Chan.
\newblock Visual {H}aystacks: {A} {V}ision-{C}entric {N}eedle-{I}n-{A}-{H}aystack {B}enchmark.
\newblock In \emph{Proceedings of the 13th International Conference on Learning Representations}, ICLR 2025, 2025.
\newblock URL \url{https://openreview.net/forum?id=9JCNPFL1f9}.

\bibitem[Huang et~al.(2025)]{Huang2025}
{Chien-yu} Huang et~al.
\newblock Dynamic-{SUPERB} {P}hase-2: {A} {C}ollaboratively {E}xpanding {B}enchmark for {M}easuring the {C}apabilities of {S}poken {L}anguage {M}odels with 180 {T}asks.
\newblock In \emph{Proceedings of the 13th International Conference on Learning Representations}, ICLR 2025, 2025.
\newblock URL \url{https://openreview.net/forum?id=s7lzZpAW7T}.

\bibitem[Shah et~al.(2025)Shah, Noguero, Heikkil{\"a}, Raj, and Kourtellis]{Shah2025}
Muhammad~A Shah, David~Solans Noguero, Mikko~A. Heikkil{\"a}, Bhiksha Raj, and Nicolas Kourtellis.
\newblock Speech {R}obust {B}ench: {A} {R}obustness {B}enchmark {F}or {S}peech {R}ecognition.
\newblock In \emph{Proceedings of the 13th International Conference on Learning Representations}, ICLR 2025, 2025.
\newblock URL \url{https://openreview.net/forum?id=D0LuQNZfEl}.

\bibitem[Chen et~al.(2025)Chen, Liu, Huang, Pei, Xu, He, Lu, Wang, and Wang]{Chen2025}
Guo Chen, Yicheng Liu, Yifei Huang, Baoqi Pei, Jilan Xu, Yuping He, Tong Lu, Yali Wang, and Limin Wang.
\newblock {CG}-{B}ench: {C}lue-grounded {Q}uestion {A}nswering {B}enchmark for {L}ong {V}ideo {U}nderstanding.
\newblock In \emph{Proceedings of the 13th International Conference on Learning Representations}, ICLR 2025, 2025.
\newblock URL \url{https://openreview.net/forum?id=le4IoZZHy1}.

\bibitem[Ye et~al.(2025)Ye, Zheng, Xue, Shen, Wang, Ma, Wang, Wang, Zhou, and Gu]{Ye2025}
Fei Ye, Zaixiang Zheng, Dongyu Xue, Yuning Shen, Lihao Wang, Yiming Ma, Yan Wang, Xinyou Wang, Xiangxin Zhou, and Quanquan Gu.
\newblock Protein{B}ench: {A} {H}olistic {E}valuation of {P}rotein {F}oundation {M}odels.
\newblock In \emph{Proceedings of the 13th International Conference on Learning Representations}, ICLR 2025, 2025.
\newblock URL \url{https://openreview.net/forum?id=BksqWM8737}.

\bibitem[Gautam et~al.(2025)Gautam, Garg, Jang, Sundaresan, and Moghaddam]{Gautam2025}
Dhruv Gautam, Spandan Garg, Jinu Jang, Neel Sundaresan, and Roshanak~Zilouchian Moghaddam.
\newblock Refactor{B}ench: {E}valuating {S}tateful {R}easoning in {L}anguage {A}gents {T}hrough {C}ode.
\newblock In \emph{Proceedings of the 13th International Conference on Learning Representations}, ICLR 2025, 2025.
\newblock URL \url{https://openreview.net/forum?id=NiNIthntx7}.

\bibitem[Jain et~al.(2025)Jain, Han, Gu, Li, Yan, Zhang, Wang, Solar-Lezama, Sen, and Stoica]{Jain2025}
Naman Jain, King Han, Alex Gu, Wen-Ding Li, Fanjia Yan, Tianjun Zhang, Sida Wang, Armando Solar-Lezama, Koushik Sen, and Ion Stoica.
\newblock Live{C}ode{B}ench: {H}olistic and {C}ontamination {F}ree {E}valuation of {L}arge {L}anguage {M}odels for {C}ode.
\newblock In \emph{Proceedings of the 13th International Conference on Learning Representations}, ICLR 2025, 2025.
\newblock URL \url{https://openreview.net/forum?id=chfJJYC3iL}.

\bibitem[Jimenez et~al.(2024)Jimenez, Yang, Wettig, Yao, Pei, Press, and Narasimhan]{Jimenez2024}
Carlos~E. Jimenez, John Yang, Alexander Wettig, Shunyu Yao, Kexin Pei, Ofir Press, and Karthik Narasimhan.
\newblock {SWE}-bench: {C}an {L}anguage {M}odels {R}esolve {R}eal-{W}orld {G}it{H}ub {I}ssues?
\newblock In \emph{Proceedings of the 12th International Conference on Learning Representations}, ICLR 2024, 2024.
\newblock URL \url{https://openreview.net/forum?id=VTF8yNQM66}.

\bibitem[Glazer et~al.(2024)Glazer, Erdil, Besiroglu, et~al.]{Glazer2024}
Elliot Glazer, Ege Erdil, Tamay Besiroglu, et~al.
\newblock Frontier{M}ath: {A} {B}enchmark for {E}valuating {A}dvanced {M}athematical {R}easoning in {AI}, 2024.
\newblock URL \url{https://arxiv.org/abs/2411.04872}.

\bibitem[Phan et~al.(2025)Phan, Gatti, Han, Li, et~al.]{Phan2025}
Long Phan, Alice Gatti, Ziwen Han, Nathaniel Li, et~al.
\newblock Humanity's {L}ast {E}xam, 2025.
\newblock URL \url{https://arxiv.org/abs/2501.14249}.

\bibitem[Chollet(2019)]{Chollet2019}
François Chollet.
\newblock On the {M}easure of {I}ntelligence, 2019.
\newblock URL \url{https://arxiv.org/abs/1911.01547}.

\bibitem[Howard et~al.(2018)Howard, Park, and Kan]{Howard2018}
Addison Howard, Eunbyung Park, and Wendy Kan.
\newblock Image{N}et {O}bject {L}ocalization {C}hallenge.
\newblock \url{https://kaggle.com/competitions/imagenet-object-localization-challenge}, 2018.
\newblock Kaggle.

\bibitem[LeGris et~al.(2024)LeGris, Vong, Lake, and Gureckis]{LeGris2024}
Solim LeGris, Wai~Keen Vong, Brenden~M. Lake, and Todd~M. Gureckis.
\newblock {H-ARC}: {A} {R}obust {E}stimate of {H}uman {P}erformance on the {A}bstraction and {R}easoning {C}orpus {B}enchmark, 2024.
\newblock URL \url{https://arxiv.org/abs/2409.01374}.

\bibitem[Opiełka et~al.(2024)Opiełka, Rosenbusch, Vijverberg, and Stevenson]{Opielka2024}
Gustaw Opiełka, Hannes Rosenbusch, Veerle Vijverberg, and Claire~E. Stevenson.
\newblock Do {L}arge {L}anguage {M}odels {S}olve {ARC} {V}isual {A}nalogies {L}ike {P}eople {D}o?, 2024.
\newblock URL \url{https://arxiv.org/abs/2403.09734}.

\bibitem[Chollet et~al.(2024{\natexlab{a}})Chollet, Knoop, Landers, Kamradt, Jud, Reade, and Howard]{arc-prize-2024}
François Chollet, Mike Knoop, Bryan Landers, Greg Kamradt, Hansueli Jud, Walter Reade, and Addison Howard.
\newblock {ARC} {P}rize 2024.
\newblock \url{https://kaggle.com/competitions/arc-prize-2024}, 2024{\natexlab{a}}.
\newblock Kaggle.

\bibitem[Chollet et~al.(2024{\natexlab{b}})Chollet, Knoop, Kamradt, and Landers]{Chollet2024}
François Chollet, Mike Knoop, Gregory Kamradt, and Bryan Landers.
\newblock {ARC} {P}rize 2024: {T}echnical {R}eport, 2024{\natexlab{b}}.
\newblock URL \url{https://arxiv.org/abs/2412.04604}.

\bibitem[Chollet et~al.(2025{\natexlab{a}})Chollet, Knoop, Kamradt, Landers, and Pinkard]{Chollet2025}
Francois Chollet, Mike Knoop, Gregory Kamradt, Bryan Landers, and Henry Pinkard.
\newblock {ARC-AGI}-2: {A} {N}ew {C}hallenge for {F}rontier {AI} {R}easoning {S}ystems, 2025{\natexlab{a}}.
\newblock URL \url{https://arxiv.org/abs/2505.11831}.

\bibitem[Chollet et~al.(2025{\natexlab{b}})Chollet, Knoop, Kamradt, Reade, and Howard]{arc-prize-2025}
François Chollet, Mike Knoop, Greg Kamradt, Walter Reade, and Addison Howard.
\newblock {ARC} {P}rize 2025.
\newblock \url{https://kaggle.com/competitions/arc-prize-2025}, 2025{\natexlab{b}}.
\newblock Kaggle.

\bibitem[Hodel(2024)]{Hodel2024}
Michael Hodel.
\newblock Addressing the {A}bstraction and {R}easoning {C}orpus via {P}rocedural {E}xample {G}eneration, 2024.
\newblock URL \url{https://arxiv.org/abs/2404.07353}.

\bibitem[Li et~al.(2025)Li, Hu, Larsen, Wu, Alford, Woo, Dunn, Tang, Zheng, Pu, and Ellis]{Li2025}
Wen-Ding Li, Keya Hu, Carter Larsen, Yuqing Wu, Simon Alford, Caleb Woo, Spencer~M. Dunn, Hao Tang, Wei-Long Zheng, Yewen Pu, and Kevin Ellis.
\newblock Combining {I}nduction and {T}ransduction for {A}bstract {R}easoning.
\newblock In \emph{Proceedings of the 13th International Conference on Learning Representations}, ICLR 2025, 2025.
\newblock URL \url{https://openreview.net/forum?id=UmdotAAVDe}.

\bibitem[Acquaviva et~al.(2022)Acquaviva, Pu, Kryven, Sechopoulos, Wong, Ecanow, Nye, Tessler, and Tenenbaum]{Acquaviva2022}
Samuel Acquaviva, Yewen Pu, Marta Kryven, Theodoros Sechopoulos, Catherine Wong, Gabrielle~E Ecanow, Maxwell Nye, Michael~Henry Tessler, and Joshua~B. Tenenbaum.
\newblock Communicating {N}atural {P}rograms to {H}umans and {M}achines.
\newblock In \emph{Proceedings of the 36th Conference on Neural Information Processing Systems}, NeurIPS 2022, 2022.
\newblock URL \url{https://dl.acm.org/doi/10.5555/3600270.3600540}.

\bibitem[Brown et~al.(2020)Brown, Mann, Ryder, Subbiah, Kaplan, Dhariwal, Neelakantan, Shyam, Sastry, Askell, Agarwal, Herbert-Voss, Krueger, Henighan, Child, Ramesh, Ziegler, Wu, Winter, Hesse, Chen, Sigler, Litwin, Gray, Chess, Clark, Berner, McCandlish, Radford, Sutskever, and Amodei]{Brown2020}
Tom~B. Brown, Benjamin Mann, Nick Ryder, Melanie Subbiah, Jared Kaplan, Prafulla Dhariwal, Arvind Neelakantan, Pranav Shyam, Girish Sastry, Amanda Askell, Sandhini Agarwal, Ariel Herbert-Voss, Gretchen Krueger, Tom Henighan, Rewon Child, Aditya Ramesh, Daniel~M. Ziegler, Jeffrey Wu, Clemens Winter, Christopher Hesse, Mark Chen, Eric Sigler, Mateusz Litwin, Scott Gray, Benjamin Chess, Jack Clark, Christopher Berner, Sam McCandlish, Alec Radford, Ilya Sutskever, and Dario Amodei.
\newblock Language {M}odels are {F}ew-{S}hot {L}earners.
\newblock In \emph{Proceedings of the 34th Conference on Neural Information Processing Systems}, NeurIPS 2020, pages 1877--1901, 2020.
\newblock URL \url{https://dl.acm.org/doi/abs/10.5555/3495724.3495883}.

\bibitem[Ravi and Larochelle(2017)]{Ravi2017}
Sachin Ravi and Hugo Larochelle.
\newblock Optimization as a {M}odel for {F}ew-{S}hot {L}earning.
\newblock In \emph{Proceedings of the 5th International Conference on Learning Representations}, ICLR 2017, 2017.
\newblock URL \url{https://openreview.net/forum?id=rJY0-Kcll}.

\bibitem[Alford et~al.(2022)Alford, Gandhi, Rangamani, Banburski, Wang, Dandekar, Chin, Poggio, and Chin]{Alford2022}
Simon Alford, Anshula Gandhi, Akshay Rangamani, Andrzej Banburski, Tony Wang, Sylee Dandekar, John Chin, Tomaso Poggio, and Peter Chin.
\newblock Neural-{G}uided, {B}idirectional {P}rogram {S}earch for {A}bstraction and {R}easoning.
\newblock In \emph{Proceedings of the 10th International Conference on Complex Networks and Their Applications}, pages 657--668, 2022.

\bibitem[Macfarlane and Bonnet(2025)]{Macfarlane2025}
Matthew Macfarlane and Cl{\'e}ment Bonnet.
\newblock Searching {L}atent {P}rogram {S}paces.
\newblock In \emph{ICML 2025 Workshop on Programmatic Representations for Agent Learning}, 2025.
\newblock URL \url{https://openreview.net/forum?id=Va937YZ9it}.

\bibitem[Ainooson et~al.(2023)Ainooson, Sanyal, Michelson, Yang, and Kunda]{Ainooson2023}
James Ainooson, Deepayan Sanyal, Joel~P. Michelson, Yuan Yang, and Maithilee Kunda.
\newblock A {N}eurodiversity-{I}nspired {S}olver for the {A}bstraction \& {R}easoning {C}orpus ({ARC}) {U}sing {V}isual {I}magery and {P}rogram {S}ynthesis, 2023.
\newblock URL \url{https://arxiv.org/abs/2302.09425}.

\bibitem[Bober-Irizar and Banerjee(2024)]{Bober2024}
Mikel Bober-Irizar and Soumya Banerjee.
\newblock Neural networks for abstraction and reasoning.
\newblock \emph{Scientific Reports}, 14, 2024.
\newblock URL \url{https://doi.org/10.1038/s41598-024-73582-7}.

\bibitem[Hocquette and Cropper(2025)]{Hocquette2025}
Céline Hocquette and Andrew Cropper.
\newblock Relational {D}ecomposition for {P}rogram {S}ynthesis, 2025.
\newblock URL \url{https://arxiv.org/abs/2408.12212}.

\bibitem[Huang et~al.(2023)Huang, Nan, Hu, Jin, Peng, Wen, Zhang, Du, Guo, Pu, and Chen]{Huang2023}
Di~Huang, Ziyuan Nan, Xing Hu, Pengwei Jin, Shaohui Peng, Yuanbo Wen, Rui Zhang, Zidong Du, Qi~Guo, Yewen Pu, and Yunji Chen.
\newblock {ANPL}: {T}owards {N}atural {P}rogramming with {I}nteractive {D}ecomposition.
\newblock In \emph{Proceedings of the 37th Conference on Neural Information Processing Systems}, NeurIPS 2023, 2023.
\newblock URL \url{https://openreview.net/forum?id=RTRS3ZTsSj}.

\bibitem[Witt et~al.(2025)Witt, Duman\v{c}i\'{c}, Guns, and Carbon]{Witt2025}
Jonas Witt, Sebastijan Duman\v{c}i\'{c}, Tias Guns, and Claus-Christian Carbon.
\newblock A {D}ivide, {A}lign, and {C}onquer {S}trategy for {P}rogram {S}ynthesis.
\newblock \emph{Journal of Artificial Intelligence Research}, 82:\penalty0 1961--1997, March 2025.
\newblock URL \url{https://doi.org/10.1613/jair.1.16847}.

\bibitem[Butt et~al.(2024)Butt, Manczak, Wiggers, Rainone, Zhang, Defferrard, and Cohen]{Butt2024}
Natasha Butt, Blazej Manczak, Auke Wiggers, Corrado Rainone, David~W. Zhang, Micha\"{e}l Defferrard, and Taco Cohen.
\newblock Code{I}t: {S}elf-{I}mproving {L}anguage {M}odels with {P}rioritized {H}indsight {R}eplay.
\newblock In \emph{Proceedings of the 41st International Conference on Machine Learning}, ICML 2024, pages 5013--5034, 2024.
\newblock URL \url{https://proceedings.mlr.press/v235/butt24a.html}.

\bibitem[Tang et~al.(2024{\natexlab{a}})Tang, Hu, Zhou, Zhong, Zheng, Si, and Ellis]{Tang2024repair}
Hao Tang, Keya Hu, Jin~Peng Zhou, Si~Cheng Zhong, Wei-Long Zheng, Xujie Si, and Kevin Ellis.
\newblock Code {R}epair with {LLM}s gives an {E}xploration-{E}xploitation {T}radeoff.
\newblock In \emph{Proceedings of the 38th Conference on Neural Information Processing Systems}, NeurIPS 2024, 2024{\natexlab{a}}.
\newblock URL \url{https://openreview.net/forum?id=o863gX6DxA}.

\bibitem[Ouellette(2024)]{Ouellette2024}
Simon Ouellette.
\newblock Towards {E}fficient {N}eurally-{G}uided {P}rogram {I}nduction for {ARC-AGI}, 2024.
\newblock URL \url{https://arxiv.org/abs/2411.17708}.

\bibitem[Wang et~al.(2024)Wang, Zelikman, Poesia, Pu, Haber, and Goodman]{Wang2024}
Ruocheng Wang, Eric Zelikman, Gabriel Poesia, Yewen Pu, Nick Haber, and Noah Goodman.
\newblock Hypothesis {S}earch: {I}nductive {R}easoning with {L}anguage {M}odels.
\newblock In \emph{Proceedings of the 12th International Conference on Learning Representations}, ICLR 2024, 2024.
\newblock URL \url{https://openreview.net/forum?id=G7UtIGQmjm}.

\bibitem[Veldkamp et~al.(2023)Veldkamp, Rosenbusch, Thoms, and Stevenson]{Veldkamp2023}
Karel Veldkamp, Hannes Rosenbusch, Luca Thoms, and Claire Stevenson.
\newblock Solving {ARC} visual analogies with neural embeddings and vector arithmetic: {A} generalized method.
\newblock In \emph{OSF}, 2023.
\newblock URL \url{https://doi.org/10.17605/OSF.IO/AKP86}.

\bibitem[Liao and Gu(2025)]{Liao2025}
Isaac Liao and Albert Gu.
\newblock {ARC-AGI} {W}ithout {P}retraining, 2025.
\newblock URL \url{https://iliao2345.github.io/blog_posts/arc_agi_without_pretraining/arc_agi_without_pretraining.html}.

\bibitem[Berman(2024)]{Berman2024}
Jeremy Berman.
\newblock How {I} came in first on {ARC-AGI-Pub} using {S}onnet 3.5 with {E}volutionary {T}est-time {C}ompute, 2024.
\newblock URL \url{https://jeremyberman.substack.com/p/how-i-got-a-record-536-on-arc-agi}.

\bibitem[Greenblatt(2024)]{Greenblatt2024}
Ryan Greenblatt.
\newblock Getting 50\% ({SoTA}) on {ARC-AGI} with {GPT}-4o, 2024.
\newblock URL \url{https://redwoodresearch.substack.com/p/getting-50-sota-on-arc-agi-with-gpt}.

\bibitem[Tan and Motani(2023)]{Tan2023}
John Chong~Min Tan and Mehul Motani.
\newblock Large {L}anguage {M}odel ({LLM}) as a {S}ystem of {M}ultiple {E}xpert {A}gents: {A}n {A}pproach to solve the {A}bstraction and {R}easoning {C}orpus ({ARC}) {C}hallenge, 2023.
\newblock URL \url{https://arxiv.org/abs/2310.05146}.

\bibitem[Fletcher-Hill(2024)]{FletcherHill2024}
Paul Fletcher-Hill.
\newblock Mini-{ARC}: {S}olving {A}bstraction and {R}easoning {P}uzzles with {S}mall {T}ransformer {M}odels, 2024.
\newblock URL \url{https://www.paulfletcherhill.com/mini-arc.pdf}.

\bibitem[Puget(2024)]{Puget2024}
Jean-Francois Puget.
\newblock A {2D} {nGPT} {M}odel {F}or {A}rc {P}rize, 2024.
\newblock URL \url{https://github.com/jfpuget/ARC-AGI-Challenge-2024/blob/main/arc.pdf}.

\bibitem[Akyürek et~al.(2025)Akyürek, Damani, Zweiger, Qiu, Guo, Pari, Kim, and Andreas]{Akyurek2025}
Ekin Akyürek, Mehul Damani, Adam Zweiger, Linlu Qiu, Han Guo, Jyothish Pari, Yoon Kim, and Jacob Andreas.
\newblock The {S}urprising {E}ffectiveness of {T}est-{T}ime {T}raining for {F}ew-{S}hot {L}earning, 2025.
\newblock URL \url{https://arxiv.org/abs/2411.07279}.

\bibitem[Lee et~al.(2025)Lee, Sim, Shin, Seo, Park, Lee, Hwang, Kim, and Kim]{Lee2025}
Seungpil Lee, Woochang Sim, Donghyeon Shin, Wongyu Seo, Jiwon Park, Seokki Lee, Sanha Hwang, Sejin Kim, and Sundong Kim.
\newblock {R}easoning {A}bilities of {L}arge {L}anguage {M}odels: {I}n-{D}epth {A}nalysis on the {A}bstraction and {R}easoning {C}orpus.
\newblock \emph{ACM Transactions on Intelligent Systems and Technology}, January 2025.
\newblock URL \url{https://doi.org/10.1145/3712701}.

\bibitem[Franzen et~al.(2024)Franzen, Disselhoff, and Hartmann]{Franzen2024}
Daniel Franzen, Jan Disselhoff, and David Hartmann.
\newblock The {LLM} {ARC}hitect: {S}olving {ARC-AGI} {I}s {A} {M}atter of {P}erspective, 2024.
\newblock URL \url{https://github.com/da-fr/arc-prize-2024/blob/main/the_architects.pdf}.

\bibitem[Moskvichev et~al.(2023)Moskvichev, Odouard, and Mitchell]{Moskvichev2023}
Arsenii~Kirillovich Moskvichev, Victor~Vikram Odouard, and Melanie Mitchell.
\newblock The {C}oncept{ARC} {B}enchmark: {E}valuating {U}nderstanding and {G}eneralization in the {ARC} {D}omain.
\newblock \emph{Transactions on Machine Learning Research}, 2023.
\newblock URL \url{https://openreview.net/forum?id=8ykyGbtt2q}.

\bibitem[Austin et~al.(2021)Austin, Odena, Nye, Bosma, Michalewski, Dohan, Jiang, Cai, Terry, Le, and Sutton]{Austin2021}
Jacob Austin, Augustus Odena, Maxwell Nye, Maarten Bosma, Henryk Michalewski, David Dohan, Ellen Jiang, Carrie Cai, Michael Terry, Quoc Le, and Charles Sutton.
\newblock Program {S}ynthesis with {L}arge {L}anguage {M}odels, 2021.
\newblock URL \url{https://arxiv.org/abs/2108.07732}.

\bibitem[Bowers et~al.(2023)Bowers, Olausson, Wong, Grand, Tenenbaum, Ellis, and Solar-Lezama]{Bowers2023}
Matthew Bowers, Theo~X. Olausson, Lionel Wong, Gabriel Grand, Joshua~B. Tenenbaum, Kevin Ellis, and Armando Solar-Lezama.
\newblock Top-{D}own {S}ynthesis for {L}ibrary {L}earning.
\newblock In \emph{Proceedings of the 50th ACM SIGPLAN Symposium on Principles of Programming Languages}, POPL 2023, pages 1182--1213, 2023.
\newblock URL \url{https://doi.org/10.1145/3571234}.

\bibitem[Chen et~al.(2021)Chen, Tworek, Jun, Yuan, de~Oliveira~Pinto, Kaplan, et~al.]{Chen2021}
Mark Chen, Jerry Tworek, Heewoo Jun, Qiming Yuan, Henrique~Ponde de~Oliveira~Pinto, Jared Kaplan, et~al.
\newblock Evaluating {L}arge {L}anguage {M}odels {T}rained on {C}ode, 2021.
\newblock URL \url{https://arxiv.org/abs/2107.03374}.

\bibitem[Devlin et~al.(2017)Devlin, Uesato, Bhupatiraju, Singh, Mohamed, and Kohli]{Devlin2017}
Jacob Devlin, Jonathan Uesato, Surya Bhupatiraju, Rishabh Singh, {Abdel-rahman} Mohamed, and Pushmeet Kohli.
\newblock {R}obust{F}ill: {N}eural {P}rogram {L}earning under {N}oisy {I}/{O}.
\newblock In \emph{Proceedings of the 34th International Conference on Machine Learning}, ICML 2017, pages 990--998, 2017.
\newblock URL \url{https://proceedings.mlr.press/v70/devlin17a.html}.

\bibitem[Du et~al.(2024)Du, Liu, Wang, Wang, Liu, Chen, Feng, Sha, Peng, and Lou]{Du2024}
Xueying Du, Mingwei Liu, Kaixin Wang, Hanlin Wang, Junwei Liu, Yixuan Chen, Jiayi Feng, Chaofeng Sha, Xin Peng, and Yiling Lou.
\newblock Evaluating {L}arge {L}anguage {M}odels in {C}lass-{L}evel {C}ode {G}eneration.
\newblock In \emph{Proceedings of the IEEE/ACM 46th International Conference on Software Engineering}, ICSE 2024, 2024.
\newblock URL \url{https://doi.org/10.1145/3597503.3639219}.

\bibitem[Ellis et~al.(2021)Ellis, Wong, Nye, Sabl\'{e}-Meyer, Morales, Hewitt, Cary, Solar-Lezama, and Tenenbaum]{Ellis2021}
Kevin Ellis, Catherine Wong, Maxwell Nye, Mathias Sabl\'{e}-Meyer, Lucas Morales, Luke Hewitt, Luc Cary, Armando Solar-Lezama, and Joshua~B. Tenenbaum.
\newblock Dream{C}oder: {B}ootstrapping {I}nductive {P}rogram {S}ynthesis with {W}ake-{S}leep {L}ibrary {L}earning.
\newblock In \emph{Proceedings of the 42nd ACM SIGPLAN International Conference on Programming Language Design and Implementation}, PLDI 2021, pages 835--850, 2021.
\newblock URL \url{https://doi.org/10.1145/3453483.3454080}.

\bibitem[Evans et~al.(2021)Evans, Bošnjak, Buesing, Ellis, Pfau, Kohli, and Sergot]{Evans2021}
Richard Evans, Matko Bošnjak, Lars Buesing, Kevin Ellis, David Pfau, Pushmeet Kohli, and Marek Sergot.
\newblock Making sense of raw input.
\newblock \emph{Artificial Intelligence}, 299:\penalty0 103521, 2021.
\newblock URL \url{https://doi.org/10.1016/j.artint.2021.103521}.

\bibitem[Fried et~al.(2023)Fried, Aghajanyan, Lin, Wang, Wallace, Shi, Zhong, Yih, Zettlemoyer, and Lewis]{Fried2023}
Daniel Fried, Armen Aghajanyan, Jessy Lin, Sida Wang, Eric Wallace, Freda Shi, Ruiqi Zhong, Scott Yih, Luke Zettlemoyer, and Mike Lewis.
\newblock In{C}oder: {A} {G}enerative {M}odel for {C}ode {I}nfilling and {S}ynthesis.
\newblock In \emph{Proceedings of the 11th International Conference on Learning Representations}, ICLR 2023, 2023.
\newblock URL \url{https://openreview.net/forum?id=hQwb-lbM6EL}.

\bibitem[Gao et~al.(2023)Gao, Madaan, Zhou, Alon, Liu, Yang, Callan, and Neubig]{Gao2023}
Luyu Gao, Aman Madaan, Shuyan Zhou, Uri Alon, Pengfei Liu, Yiming Yang, Jamie Callan, and Graham Neubig.
\newblock {PAL}: {P}rogram-aided {L}anguage {M}odels.
\newblock In \emph{Proceedings of the 40th International Conference on Machine Learning}, ICML 2023, pages 10764--10799, 2023.
\newblock URL \url{https://proceedings.mlr.press/v202/gao23f.html}.

\bibitem[Li et~al.(2024)Li, Liang, Zeng, Chen, Hausman, Sadigh, Levine, Fei-Fei, Xia, and Ichter]{Li2024icml}
Chengshu Li, Jacky Liang, Andy Zeng, Xinyun Chen, Karol Hausman, Dorsa Sadigh, Sergey Levine, Li~Fei-Fei, Fei Xia, and Brian Ichter.
\newblock Chain of {C}ode: {R}easoning with a {L}anguage {M}odel-{A}ugmented {C}ode {E}mulator.
\newblock In \emph{Proceedings of the 41st International Conference on Machine Learning}, ICML 2024, pages 28259--28277, 2024.
\newblock URL \url{https://proceedings.mlr.press/v235/li24ar.html}.

\bibitem[Li and Ellis(2024)]{Li2024}
Wen-Ding Li and Kevin Ellis.
\newblock Is {P}rogramming by {E}xample {S}olved by {LLM}s?
\newblock In \emph{Proceedings of the 38th Conference on Neural Information Processing Systems}, NeurIPS 2024, 2024.
\newblock URL \url{https://openreview.net/forum?id=xqc8yyhScL}.

\bibitem[Li et~al.(2022)Li, Choi, Chung, Kushman, Schrittwieser, Leblond, Eccles, Keeling, Gimeno, Lago, Hubert, Choy, de~Masson~d’Autume, Babuschkin, Chen, Huang, Welbl, Gowal, Cherepanov, Molloy, Mankowitz, Robson, Kohli, de~Freitas, Kavukcuoglu, and Vinyals]{Li2022}
Yujia Li, David Choi, Junyoung Chung, Nate Kushman, Julian Schrittwieser, Rémi Leblond, Tom Eccles, James Keeling, Felix Gimeno, Agustin~Dal Lago, Thomas Hubert, Peter Choy, Cyprien de~Masson~d’Autume, Igor Babuschkin, Xinyun Chen, Po-Sen Huang, Johannes Welbl, Sven Gowal, Alexey Cherepanov, James Molloy, Daniel~J. Mankowitz, Esme~Sutherland Robson, Pushmeet Kohli, Nando de~Freitas, Koray Kavukcuoglu, and Oriol Vinyals.
\newblock Competition-{L}evel {C}ode {G}eneration with {A}lpha{C}ode.
\newblock \emph{Science}, 378\penalty0 (6624):\penalty0 1092--1097, 2022.
\newblock URL \url{https://www.science.org/doi/abs/10.1126/science.abq1158}.

\bibitem[Nijkamp et~al.(2023)Nijkamp, Pang, Hayashi, Tu, Wang, Zhou, Savarese, and Xiong]{Nijkamp2023}
Erik Nijkamp, Bo~Pang, Hiroaki Hayashi, Lifu Tu, Huan Wang, Yingbo Zhou, Silvio Savarese, and Caiming Xiong.
\newblock Code{G}en: {A}n {O}pen {L}arge {L}anguage {M}odel for {C}ode with {M}ulti-{T}urn {P}rogram {S}ynthesis.
\newblock In \emph{Proceedings of the 11th International Conference on Learning Representations}, ICLR 2023, 2023.
\newblock URL \url{https://openreview.net/forum?id=iaYcJKpY2B_}.

\bibitem[Qiu et~al.(2025)Qiu, Zeng, Ezick, Lott, and Tong]{Qiu2025}
Ruizhong Qiu, Weiliang~Will Zeng, James Ezick, Christopher Lott, and Hanghang Tong.
\newblock How efficient is {LLM}-generated code? {A} rigorous \& high-standard benchmark.
\newblock In \emph{Proceedings of the 13th International Conference on Learning Representations}, ICLR 2025, 2025.
\newblock URL \url{https://openreview.net/forum?id=suz4utPr9Y}.

\bibitem[Schmidhuber(2004)]{Schmidhuber2004}
J\"{u}rgen Schmidhuber.
\newblock Optimal {O}rdered {P}roblem {S}olver.
\newblock \emph{Machine Learning}, 54\penalty0 (3):\penalty0 211--254, March 2004.
\newblock URL \url{https://doi.org/10.1023/B:MACH.0000015880.99707.b2}.

\bibitem[Tang and Ellis(2023)]{Tang2023}
Hao Tang and Kevin Ellis.
\newblock From {P}erception to {P}rograms: {R}egularize, {O}verparameterize, and {A}mortize.
\newblock In \emph{Proceedings of the 40th International Conference on Machine Learning}, ICML 2023, pages 33616--33631, 2023.
\newblock URL \url{https://proceedings.mlr.press/v202/tang23c.html}.

\bibitem[Tang et~al.(2024{\natexlab{b}})Tang, Key, and Ellis]{Tang2024worldcoder}
Hao Tang, Darren~Yan Key, and Kevin Ellis.
\newblock World{C}oder, a {M}odel-{B}ased {LLM} {A}gent: {B}uilding {W}orld {M}odels by {W}riting {C}ode and {I}nteracting with the {E}nvironment.
\newblock In \emph{Proceedings of the 38th Conference on Neural Information Processing Systems}, NeurIPS 2024, 2024{\natexlab{b}}.
\newblock URL \url{https://openreview.net/forum?id=QGJSXMhVaL}.

\bibitem[Das et~al.(2023)Das, Tenenbaum, Solar-Lezama, and Tavares]{Das2023}
Ria Das, Joshua~B. Tenenbaum, Armando Solar-Lezama, and Zenna Tavares.
\newblock Combining {F}unctional and {A}utomata {S}ynthesis to {D}iscover {C}ausal {R}eactive {P}rograms.
\newblock In \emph{Proceedings of the 50th ACM SIGPLAN Symposium on Principles of Programming Languages}, POPL 2023, pages 1628--1658, 2023.
\newblock URL \url{https://doi.org/10.1145/3571249}.

\end{thebibliography}

\newpage

\appendix

\section{Appendix: Reproducing the \{Generator x Program\} Success Rates}

To reproduce the success rates in Table \ref{tab:comparison}, first download our ARC-GEN-100K dataset and clone various GitHub repositories as follows:\\

\mdfsetup{%
middlelinecolor=gray,
middlelinewidth=1pt,
backgroundcolor=gray!5,
roundcorner=3pt}
\begin{mdframed}
\scriptsize
\begin{alltt}
\$ curl -L -o ARC-GEN-100K.zip \textbackslash
  https://www.kaggle.com/api/v1/datasets/download/arcgen100k/the-arc-gen-100k-dataset
\$ git clone --recurse-submodules https://github.com/google/ARC-GEN.git
\$ git clone https://github.com/michaelhodel/re-arc.git
\$ git clone https://github.com/xu3kev/BARC.git
\end{alltt}
\end{mdframed}

Then, extract the evaluation utilities and unzip all example benchmarks:\\

\mdfsetup{%
middlelinecolor=gray,
middlelinewidth=1pt,
backgroundcolor=gray!5,
roundcorner=3pt}
\begin{mdframed}
\footnotesize
\begin{alltt}
\$ cp -r ARC-GEN/misc/evaluation/* . && mkdir examples && cd examples
\$ mkdir ARC-GEN-100K && cd ARC-GEN-100K && unzip ../../ARC-GEN-100K.zip && cd ..
\$ unzip ../BARC.zip && unzip ../re-arc/re_arc.zip && cd ..
\end{alltt}
\end{mdframed}

Evaluation using the RE-ARC verifiers should be done from the \texttt{re-arc} directory, e.g.:\\

\mdfsetup{%
middlelinecolor=gray,
middlelinewidth=1pt,
backgroundcolor=gray!5,
roundcorner=3pt}
\begin{mdframed}
\footnotesize
\begin{alltt}
\$ cd re-arc

\$ python3 evaluate_using_re_arc.py ../examples/BARC/tasks
\textcolor{darkgray}{Testing task 007bbfb7 ... pass
Testing task 00d62c1b ... pass
Testing task 017c7c7b ... FAIL
[...]
Examples pass for 65/400 tasks (\textcolor{red}{16.25\%})}

\$ python3 evaluate_using_re_arc.py ../examples/ARC-GEN-100K
\textcolor{darkgray}{Testing task 007bbfb7 ... pass
Testing task 00d62c1b ... pass
[...]
Examples pass for 400/400 tasks (\textcolor{ForestGreen}{100\%})}

\$ cd ..
\end{alltt}
\end{mdframed}

Evaluation using the BARC source programs should be done from the \texttt{BARC} directory, e.g.:\\

\mdfsetup{%
middlelinecolor=gray,
middlelinewidth=1pt,
backgroundcolor=gray!5,
roundcorner=3pt}
\begin{mdframed}
\footnotesize
\begin{alltt}
\$ cd BARC

\$ python3 -O evaluate_using_barc.py --exampledir=../examples/re_arc/tasks
\textcolor{darkgray}{Testing task 007bbfb7 ... pass
Testing task 00d62c1b ... FAIL
[...]
Examples pass for 4/108 tasks (\textcolor{red}{3.7\%})}

\$ python3 -O evaluate_using_barc.py --exampledir=../examples/ARC-GEN-100K
\textcolor{darkgray}{Testing task 007bbfb7 ... pass
Testing task 00d62c1b ... pass
[...]
Examples pass for 108/108 tasks (\textcolor{ForestGreen}{100\%})}

\$ cd ..
\end{alltt}
\end{mdframed}

We recommend performing these experiments in a standard Linux environment, but there are no specific hardware requirements (e.g., CPU speed, number of cores, RAM, etc).  Please allocate up to six hours of runtime for all verifiers to complete.

\end{document}